\documentclass[conference,letterpaper]{IEEEtran}
\IEEEoverridecommandlockouts
\usepackage{amsmath,amssymb,booktabs,graphicx,array,tabularx,cite}
\usepackage[hidelinks]{hyperref}
\hypersetup{pdfauthor={Ziyang Huang and Songwei Wu},pdftitle={Testing Quantized KV-Cache Maintenance: Encoding Layout and Historical Restoration}}
\title{Testing Quantized KV-Cache Maintenance:\\ Encoding Layout and Historical Restoration}
\author{\IEEEauthorblockN{Ziyang Huang and Songwei Wu}
\IEEEauthorblockA{Harbin Institute of Technology}
\thanks{Ziyang Huang and Songwei Wu contributed equally. Ziyang Huang is the corresponding author.}\thanks{This work has been submitted to the IEEE for possible publication. Copyright may be transferred without notice, after which this version may no longer be accessible.}}
\begin{document}
\bstctlcite{paperrefs:etal}
\maketitle
\begin{abstract}
Repeating rows then selecting originals retains layout-dependent drift despite
equal encoding counts. We measure when existing checks fail to establish
preservation contracts in fixed quantized KV-cache implementations.
In four-bit HQQ axis zero at sixteen cycles,
median paired relative-L2 increases over same-shape re-encoding are 7.22
percentage points for Qwen3-1.7B and 11.59 for SmolLM3-3B.
Matched encoded states cover sixteen shared questions per model
at one and sixteen cycles. Axis one matches the shape-preserving control;
lazy reorders preserve the initial state. On one known input, fixed-update
counts retain the difference; preserving group members removes it.
Our prefix-view extension passes current-prefix checks yet changes history
and next logits in six model/codec settings at residual capacities four and 128.
Public lookup reaches this boundary, including one of four
precommitted project-new GPU inputs; fixed-query swaps isolate
a conditional old-history effect.
Checkpoint/dense rollback is exact. An explicit delayed lookup client admits
a supported 129-token path excluded by native AR on one known context.
Failed task-harm expansion precludes population failure-rate or general
answer-degradation claims. Native MLX-LM fixtures expose stale caller caches;
synchronization restores the tested in-place contract with native computations
preserved. In rejection tests, stale callers change logits but not greedy tokens.
\end{abstract}

\begin{IEEEkeywords}KV cache, quantization, software testing, stateful interfaces, numerical reproducibility\end{IEEEkeywords}
\section{Introduction}
KV-cache quantization reduces attention-history storage~\cite{liu2024kivi}.
Decoding also reorders cache rows, appends residuals, flushes storage and discards
rejected tokens. Our first prefix-view repair passed all six public-generation
conditions yet failed three composed-operation tests. A pass must identify
the maintained state it validates.

Two numerical boundaries motivate the study. Repeating rows then selecting
the originals is a logical identity, yet can change quantization groups and
error. A correct crop can retain flush-altered values rather than pre-branch
history.

Prior work covers HQQ grouping/separation~\cite{jwa2026sharedstopping},
requantization~\cite{jang2025blockdialect}, NVFP4 repeated QDQ~\cite{nvidia2026kvcompression},
public regressions~\cite{transformers45510,transformerscurrenttests}
and buffered rollback~\cite{tiwari2025quantspec}. We measure when passing existing
regression or consumer checks is insufficient for a stated maintenance contract,
using matched encoding budgets to control the numerical comparison.

\textbf{RQ1:} Do equal encoding counts remove drift
differences between logical identity paths? \textbf{RQ2:} When does a repair
satisfy a current-state consumer, and when does it restore pre-branch history?
\textbf{RQ3:} Which effects reach public decoding or a completed answer?

Same-state, equal-count paths isolate extra HQQ axis-zero drift across two
models and sixteen shared questions; fixed-update/group controls test its
mechanism. Append--flush--crop distinguishes current-prefix correctness from
history restoration, including default capacity 128. Public lookup in our
extension, fixed-query swaps and an explicit delayed client's supported
129-token path connect history to sampling. Exact checkpoint/dense controls,
eleven zero-event GPU streams and favorable native removal bound the results.

Table~\ref{tab:decisions} maps insufficient acceptance evidence to each contract.
Alongside history, native MLX-LM fixtures test caller-visible updates and a
scoped synchronization repair (Section~\ref{sec:mlx}).
Reset checks have no recall advantage over the official combined regression;
mixed sampled outcomes fail the expansion rule.

\section{Study design and references}
\subsection{Objects, implementations and inputs}
The main executions used Transformers 5.20.2, PyTorch 2.10.0+cu128, the fixed
HQQ 0.2.8.post1 source and Quanto 0.2.7~\cite{transformers5202,hqq028,quanto027}.
Qwen3-1.7B and SmolLM3-3B weights and tokenizers were pinned by revision and file
hash~\cite{arxiv250509388,bakouch2025smollm3,qwen3pinnedcard,smolpinnedcard}.
GPU collections ran sequentially on one RTX 5080. CPU checks hid CUDA before
import; performance is unmeasured. Full-model cache experiments
retained native KV heads and layers; a separate codec comparison selected three
related layers and is identified as such.

For the native numerical study, we fixed the first sixteen questions of the
official GSM8K test file~\cite{arxiv211014168}, pinned to a source commit before
maintenance outcomes were observed. Their hashes differed from the previously
recorded questions. This did not establish exclusion from pretraining. Each
model produced a raw BF16 greedy prefix of 64 generated tokens per question.
We used these prefixes as fixed state-construction inputs, not as a standard
quality evaluation: Qwen3's model card advises against greedy thinking-mode
decoding. The later sampled pilot used the next two test rows, 16 and 17,
model-card temperature/top-$p$, fixed top-$k$ and separately frozen random streams.

Other checks used already observed development inputs. The codec comparison
used the first question and three layers; full-model batch checks used one or
two of the original questions. Public assisted-decoding controls used a small
random model and synthetic prompts; a later Qwen3-1.7B study used the already
observed first question and separately frozen sampling streams. A precommitted
GPU cohort then used four project-new inputs, source rows 18--21. These are different evidence units.
They are not independent task observations.

\subsection{The encoded-state reference}
Let $C=(q,\theta,r,\pi)$ denote packed codes, decoding metadata, the residual
buffer, and an optional pending batch map. Its current logical decoded state is
\begin{equation}
 V(C)=\operatorname{concat}\bigl(P_\pi D_\theta(q),r\bigr),
\end{equation}
with an absent residual omitted. In the tested lazy implementation, the pending
map applies to the packed part while residual rows are changed immediately.
This model describes the pinned implementation, not every cache format.

For a pure batch-row operation with mathematical gather $P$, let $C$ be its
input cache and $z$ the new K/V supplied to the next update. The
row-transformation contract requires that update to return
\begin{equation}
 \operatorname{concat}\bigl(PV(C),z\bigr).
 \label{eq:consumer}
\end{equation}
We distinguish returned K/V from next-model logits. A current-consumer reference
instead decodes the state after maintenance, before the next update: it checks
consumption, not whether maintenance formed the required state. Historical
references use the saved pre-branch state. Flushing may re-encode storage after
forming the return; this later storage is not the reference for that return.

Batch-repair references use DynamicCache populated with the same decoded
K/V after dense gather, then identical next tokens, masks, positions and batch
shape. An uncompressed original would include initial quantization error.

\subsection{Maintenance compositions and numerical metrics}
For the numerical study we wrote the decoded initial encoded state as $x_0$.
Each same-shape cycle decoded and encoded twice without changing shape.
Each repeat-select cycle duplicated the batch, encoded and decoded it, selected
the original rows, and encoded again. Selection after duplication was logically
the identity. Both paths therefore performed two encodings per cycle, while
the shapes seen by the first encoder differed. The third path used existing
public lazy reorders whose composition was the identity and added no encoding.
The main matrix modeled the unmerged proposal. A CPU check re-encoded retained
layer-zero K/V from each model's first question and ran its two original batch
methods in current containers. With one residual token their K/V matched the modeled operations at one and sixteen
cycles; empty-residual selection raised. This was not a complete historical
installation or deployed repair.

For concatenated all-layer K/V, we measured
\begin{equation}
 E_T=\frac{\lVert x_T-x_0\rVert_2}{\lVert x_0\rVert_2}.
\end{equation}
Key and value errors were also retained separately. Consumer distributions used
total variation, $\tfrac12\sum_i|p_i-p_{0i}|$, computed from saved logits in
float64. A maximizer changed only if the original maximizing token no longer
belonged to the changed maximum set. This avoids counting a different choice
inside a rounded tie as a strict decision change.

The native matrix covered two models, sixteen shared questions, four/eight bits
and axes zero/one: 128 configurations. Each had three paths and observations at
cycles one and sixteen, giving 768 observations. Each condition was collected
once with fixed inputs and decoding; across-seed uncertainty is unmeasured.
All question-level points remain. Medians, ranges and paired directions are
descriptive; layers, paths and settings are not independent tasks.
The chosen question order supports neither population rates nor a significance
test.

\subsection{Execution and audit boundaries}
Inputs, source hashes and protocols were fixed before each new collection.
Raw states, logits, tokens, traces, failures and truncations remain.
Independent NumPy readers recomputed saved array relationships
without importing the cache implementation. Such checks do not independently
re-execute every model. Uninstrumented generation controls had to reproduce
each output or exception. Setup failures and corrections were retained
separately from library defects.

\section{RQ1: Encoding layout in identity maintenance}
\subsection{Equal encoding counts do not remove axis-zero drift}
\begin{figure*}[t]
\centering
\includegraphics[width=\textwidth]{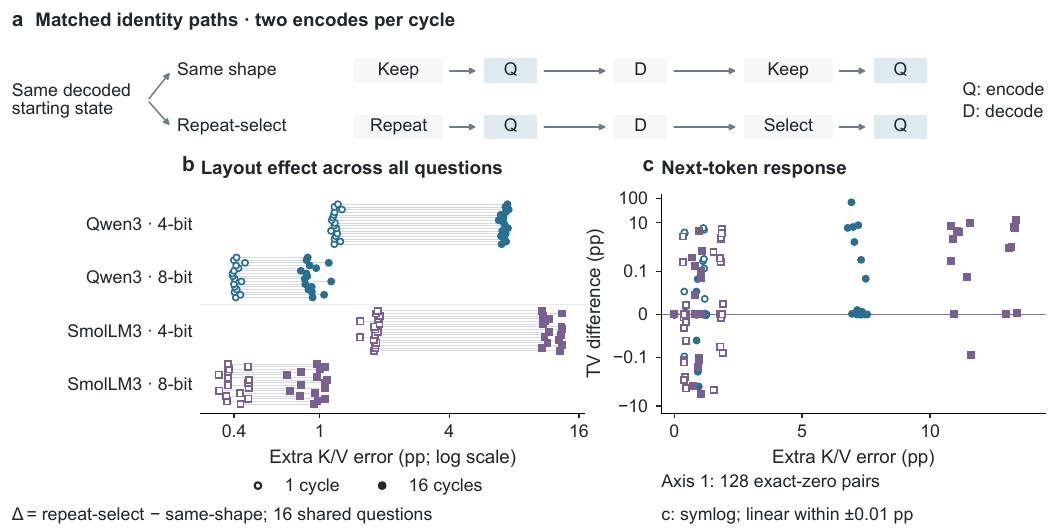}
\caption{Equal encoding counts retain layout-dependent drift. (a) Same decoded starting state; Q/D: encode/decode; two encodes per cycle. (b) All 128 axis-zero observations on one log scale; lines pair questions at one/sixteen cycles. Values are paired repeat-select minus same-shape relative-L2 error (pp). (c) All 256 state/next-token TV differences, including 128 exact-zero axis-one observations. Blue circles/purple squares: Qwen3/SmolLM3; hollow/filled: one/sixteen cycles. TV is symmetric-log, linear within 0.01 pp. All sixteen questions remain; task-ID offsets are orthogonal to measurements. Fixed related conditions have no population confidence interval.}
\label{fig:maintenance}
\end{figure*}

In all sixteen questions within each model, the four-bit axis-zero repeat-select
path produced more drift than the same-shape control at both recorded cycle
counts. Table~\ref{tab:drift} reports medians; Fig.~\ref{fig:maintenance} retains
the individual question observations and the axis-one controls. After one cycle,
Qwen3's median drift changed from 0.162\% in the control to 1.339\% in
repeat-select; SmolLM3 changed from 0.184\% to 1.997\%. Thus the numerical
effect was not restricted to sixteen-cycle stress.

\begin{table}[t]
\centering\small
\caption{Median KV drift (\%) from the same encoded starting state: HQQ axis zero, sixteen shared questions, two encodings per cycle. Same: same-shape; Repeat: repeat-select.}
\label{tab:drift}
\renewcommand{\arraystretch}{1.08}
\begin{tabular*}{\columnwidth}{@{\extracolsep{\fill}}llrrrr@{}}
\toprule
& & \multicolumn{2}{c}{One cycle} & \multicolumn{2}{c}{Sixteen cycles}\\
\cmidrule(lr){3-4}\cmidrule(lr){5-6}
Model & Bits & Same & Repeat & Same & Repeat\\
\midrule
Qwen3 & 4 & 0.162 & 1.339 & 0.216 & 7.451\\
Qwen3 & 8 & 0.197 & 0.601 & 0.446 & 1.348\\
\addlinespace[2pt]
SmolLM3 & 4 & 0.184 & 1.997 & 0.225 & 11.821\\
SmolLM3 & 8 & 0.205 & 0.600 & 0.561 & 1.486\\
\bottomrule
\end{tabular*}
\end{table}

After sixteen cycles, the corresponding medians were 0.216\% versus 7.451\%
for Qwen3, and 0.225\% versus 11.821\% for SmolLM3. Eight-bit axis-zero
conditions retained the same paired direction on all sixteen questions per
model at both cycle counts. These observations describe the fixed inputs;
they are not population defect proportions.

\subsection{Separating grouping from shared stopping}
We tested whether equal optimizer-update counts remove the
shape effect. The original diagnostic used the already-seen first question, both
models, four-bit HQQ and both axes. It reused all original decoded K/V layers,
without another model forward. Native same-shape and repeat-select outputs had
to reproduce saved one-cycle GPU arrays exactly before expansion.

We crossed two stopping rules with three layouts: same-shape, native
repeat-select, and group-preserving repeat-select. The fixed rule replaced only
the legacy solver's stopping statement, retaining its arithmetic and twenty
updates. Actual update counts were traced. For grouping, HQQ axis zero reshapes
$N=64M$ elements into $[64,M]$; duplicating a flat buffer changes the column
stride to $2M$ and therefore changes group members. An invertible permutation
instead presents the two original copies side by side within each row of
$[64,2M]$. We verified original flat indices group by group, then inverted the
permutation before selecting rows. Axis one's contiguous groups of 64 are
already preserved by duplicating the full buffer.

\begin{table}[t]
\centering\small
\caption{One-cycle repeat-to-same-shape distance $\Delta$ (\%). One known shared question, four-bit HQQ, all K/V layers. Fixed: twenty legacy updates; + Group: also preserve group members. Zeros are exact.}
\label{tab:mechanism}
\renewcommand{\arraystretch}{1.08}
\begin{tabular*}{\columnwidth}{@{\extracolsep{\fill}}llrrr@{}}
\toprule
Model & Axis & Native & Fixed & Fixed + Group\\
\midrule
Qwen3 & 0 & 1.328 & 1.331 & 0\\
Qwen3 & 1 & 0 & 0 & 0\\
\addlinespace[2pt]
SmolLM3 & 0 & 1.966 & 1.972 & 0\\
SmolLM3 & 1 & 0 & 0 & 0\\
\bottomrule
\end{tabular*}
\end{table}

Table~\ref{tab:mechanism} measures the distance between the final repeat and
same-shape states under the same stopping rule,
\mbox{$\Delta=\lVert x_{\mathrm{repeat}}-x_{\mathrm{same}}\rVert_2/
\lVert x_{\mathrm{same}}\rVert_2$}, not their errors from $x_0$.
All 256 K/V conditions reproduced both native baselines exactly. Across 1,536
paths, fixing twenty updates did not eliminate axis-zero differences:
128 of its 128 tensors remained unequal. With both fixed updates and
preserved group members, all 256 tensors matched their same-shape controls
exactly. Keeping group members also matched native same-shape outputs exactly in all 256 K/V tensors.
Iteration count alone was insufficient here. The separate Mechanism
Confirmation note checks additional fixed contexts. These observations concern
the tested layouts; they do not establish a new grouping mechanism, a general
floating-point invariant or a production repair.

\subsection{Positive controls and backend scope}
Axis one, the request-local setting in the controlled HQQ
study~\cite{jwa2026sharedstopping}, showed no extra
repeat-select effect in this matrix. Its KV and logits files were byte-identical
to the matched same-shape files in all 128 comparisons over models, questions,
bit widths and cycle counts. This equality does not mean that re-encoding itself
was exact relative to $x_0$, nor does it establish independence from arbitrary
companion requests: that is a different contrast~\cite{jwa2026sharedstopping}.
All 256 lazy-reorder observations preserved the
initial decoded state, packed representation and consumer logits under their
recorded exact controls.

A separate CPU comparison used original unquantized BF16 K/V from the same
development question in both pretrained models, cast exactly to FP32 for each
codec. Three fixed layers per model supplied related states. HQQ axes zero/one
and Quanto axis zero produced eighteen configurations and 108 maintenance
observations. The default Quanto setting had zero measured extra drift in all
tested paths and cycle counts. HQQ axis-one matched paths were equal, and HQQ
axis-zero repeat-select drift exceeded its control in all three related layers
per model at both cycles. The CPU source, dtype and selected-layer scope differ
from the native GPU collection, so these are not quality or speed rankings
between backends. Detailed source data retain all settings.

\subsection{Consumer changes are heterogeneous}
For four-bit axis-zero repeat-select at sixteen cycles, next-consumer TV had a
median of 0.00184\% and maximum 69.15\% in Qwen3. SmolLM3's median was
1.661\% and maximum 14.94\%. Only one of the 768 related observations removed
the original maximizer from the new maximum set: Qwen3, question 11, four bits,
axis zero, sixteen repeat-select cycles. It was a next-step decision effect,
not evidence that a final answer became incorrect.

The numerical matrix identifies a shape-related component under these
settings. The single-input interventions further distinguish group membership
from update count, without assigning additive causal shares to the two-stage
errors or extrapolating to all reduction geometries. Existing analyses of HQQ
batch dependence remain the nearest mechanistic reference~\cite{jwa2026sharedstopping}.

\section{RQ2: Current-state repairs and historical restoration}
\subsection{Physical fields and logical consumers}
The tested upstream lazy reorder intentionally leaves the packed fields in
their old order until the next update. Its source explicitly motivates the
deferral by avoiding a lossy quantization round trip. The fixed upstream tests
also check returned K/V after actual updates and residual
flushes~\cite{transformerscurrenttests}. Consumer-boundary checking is
established practice.

The batch repeat and selection methods did not provide the same handling of
packed state in the tested release. This defect family was already described
in the public proposal~\cite{transformers45510}. We retained it as a known
control and implemented a scoped adapter that delegates those two methods to
the existing lazy reorder while tracking logical batch size. It creates fresh
layers within the real QuantizedCache container; it does not modify the
installed library, quantizer, crop or reset.

\subsection{Retrospective lifecycle controls}
Two public fixes ground the contract checks~\cite{transformers48700,transformers48809}.
Across the two fix/parent pairs we froze 48 CPU FP32 conditions:
three seeds, Dynamic/Quanto reset cases and four Quanto beam schedules.
Six invocations of the official added regression methods retained their
function bodies with CPU/availability adaptation. Before the beam fix, six
conditions raised and six silently returned the wrong rows; all twelve fixed
conditions passed. The earlier reset pair failed six parent conditions and
passed six fixed ones. All 42 completed consumers matched references built
from the post-operation state; twelve nevertheless violated the expected
maintenance transformation.

The later quantized-reset pair supplies a negative control: all six parent and
six fixed conditions passed fresh-instance equivalence, despite a packed
object remaining in three parent conditions. Its added official reset method
also passed the parent: inherited dynamic reset already forced
reinitialization. Physical residue alone therefore did not establish a fault.

\begin{table}[t]
\caption{Four matched reset--generate settings, not independent tasks.
Cells give parent/fixed values. Len: length just after reset.
Tokens: six-token equality to a fresh cache; State: exact K/V, logits and
generation scores. $\checkmark$: match; $\times$: differ.}
\label{tab:reset_client}
\centering
\renewcommand{\arraystretch}{1.08}
\begin{tabular*}{\columnwidth}{@{\extracolsep{\fill}}llccc@{}}
\toprule
& & \multicolumn{3}{c}{Parent / fixed}\\
\cmidrule(l){3-5}
Model & Cache & Len & Tokens & State \\
\midrule
Qwen3 & Dynamic & 10/0 & $\checkmark/\checkmark$ & $\times/\checkmark$ \\
Qwen3 & Quanto & 0/0 & $\times/\checkmark$ & $\times/\checkmark$ \\
\addlinespace[2pt]
SmolLM3 & Dynamic & 10/0 & $\checkmark/\checkmark$ & $\times/\checkmark$ \\
SmolLM3 & Quanto & 0/0 & $\times/\checkmark$ & $\times/\checkmark$ \\
\bottomrule
\end{tabular*}
\end{table}

After an eight-condition random-model preflight, public \texttt{generate}
checked both pretrained models, Dynamic/Quanto caches and the earlier reset
pair. Fresh subprocesses loaded pinned cache modules before model imports;
generator/model/dependencies stayed current, not full historical installations.
Two official regression prompts (capital of France, then a number sequence
after reset), greedy generation and a six-token cap yielded 48 calls including
paired unobserved repeats; all 144 observed forwards succeeded.

All four parent settings returned text but violated fresh-instance equivalence
(Table~\ref{tab:reset_client}). Token equality passed in both Dynamic settings;
zero-length checks passed in both Quanto settings. All four fixed settings
matched fresh-cache scores, per-layer K/V and logits exactly, and every
observer control passed. The official full-model regression already checks
length and decoded output. Together these checks detect all four parent
settings here; our results establish neither new defect discovery nor a
recall advantage over that regression.

Saved first-forward traces diagnose the parents before token feedback:
Dynamic processed seven of seventeen prompt tokens, while Quanto processed
seventeen but returned twenty-seven K/V positions, ten retained. Fixed caches
processed/returned seventeen. Exact layerwise snapshot reconstruction verified
these concatenations; this post-result diagnosis adds no samples.

\subsection{Random cache conditions}
The matched FP32 matrix used batch two, two heads, dimension 64, a four-token
prefill and R4. Zero/one/three/four additional tokens covered residual and flush
stages. HQQ four-bit axes zero/one and Quanto four-bit axis zero crossed seven
traces of row repetition, selection, composition and identity cycles.

Across 84 conditions per implementation, initial states and next inputs
matched. The adapter satisfied Eq.~\ref{eq:consumer} in all 84, preserving
the original's 18 successes; the original raised 36 operation and 30 consumer
exceptions. Packed objects stayed until consumption. These are known-family
test conditions, not 66 new defects.

\subsection{Complete pretrained model integration}
One known question across two models, three backend/axis settings and three
batch traces gave eighteen full-model CPU BF16 conditions.
All consumer K/V and final logits exactly matched the same-decoded-state
DynamicCache reference.

Because repeating one row does not establish selection between different
requests, a second check used two distinct existing prefixes with unequal valid
lengths. Left padding, position IDs, masks and next tokens were gathered with
the same mathematical row map as the cache. Four traces covered repetition,
selection, composed mixed-row selection and sixteen identity cycles. All 24
additional conditions exactly matched the reference.
Source-token audits reconstructed request state; a corrected pre-forward setup
failure remains recorded separately.

These results support this adapter's tested batch-operation integration. They
do not establish arbitrary-trace correctness, answer improvement or GPU speed.
Views retain physical storage until flushing; immediate memory release is
not established.

\subsection{A prefix-view crop repair and its history boundary}
The public failure in RQ3 motivated a separate engineering extension. We added
a visible packed-prefix length to the batch-view adapter. Crop changes this
length and trims the residual, without decoding or re-encoding the packed
object at the operation boundary. The next update decodes the existing object,
applies the prefix view and pending row map, and follows the native update and
flush policy. This restores a prefix-of-current-state contract, not a snapshot
of the cache before a speculative branch.

Three backend/axis settings, four residual stages and eight crop/row-operation
traces gave 96 matched original/adapted pairs (192 frozen CPU conditions).
Each trace had one/four/one-token consumers, including a flush. The original
passed 12/96; the final adapter passed 96/96 with unchanged packed objects and
no operation-time encoding. Independent readers checked the matched initial
arrays and consumer relations. These remain known-family conditions.

A first repair had completed all six public-generation conditions but failed
three composition conditions in this matrix: the empty residual's batch shape
did not follow subsequent row maps. Crop left a rank-four empty residual that
the inherited reorder skipped. The correction uses the native rank-one empty
representation. This defect in our initial repair and its unchanged gate remain
recorded; the three failures are not newly discovered upstream bugs.

Full-model CPU BF16 integration then checked eighteen conditions across the
two pretrained models and all three settings. Packed-prefix crop,
residual-crossing crop and complete removal each had three subsequent
consumers. All 54 consumer comparisons had exact native-layer K/V and final
logits against a DynamicCache populated from the same current decoded state.
Original and adapted prefill logits were independently compared;
initial decoded-state equality was asserted during collection, but the two
initial arrays were not separately retained for independent replay.

Preserving the current prefix is weaker than restoring earlier history.
QuantSpec already addresses quantized speculative rollback with buffers that
defer quantization until verification~\cite{tiwari2025quantspec}; avoiding
unnecessary round trips is not a new principle introduced here. The following
experiment tests the stronger history contract with an actual crop and next
model call. Earlier saved-array audits remain in the research record.

\subsection{Actual rollback and the next consumer}
We next tested whether a current-prefix repair restores the state preceding
a branch. Six fixed conditions combined the same known question in both models
with four-bit HQQ axes zero/one and Quanto axis zero. Group size was 64 and
residual capacity four. Native and adapted prefill and committed-token K/V and
logits matched exactly. Each complete checkpoint retained packed codes,
quantization metadata, residual tensors, pending row maps and logical views;
JSON and plain safetensors reconstruction preserved all fields and values,
without shared tensor storage between source and restored copies.

Four paths shared that checkpoint: direct continuation, a one-token branch
cropped before flushing, a four-token branch cropped after native flushing,
and the same four-token branch followed by full-checkpoint restoration. Every
path then consumed the same forced token, attention mask and positions. A
DynamicCache populated from each path's current decoded state supplied its
current-state reference. Direct continuation supplied the historical reference.
The first condition passed independent controls before the other five ran.

\begin{figure*}[t]
\centering
\includegraphics[width=\textwidth]{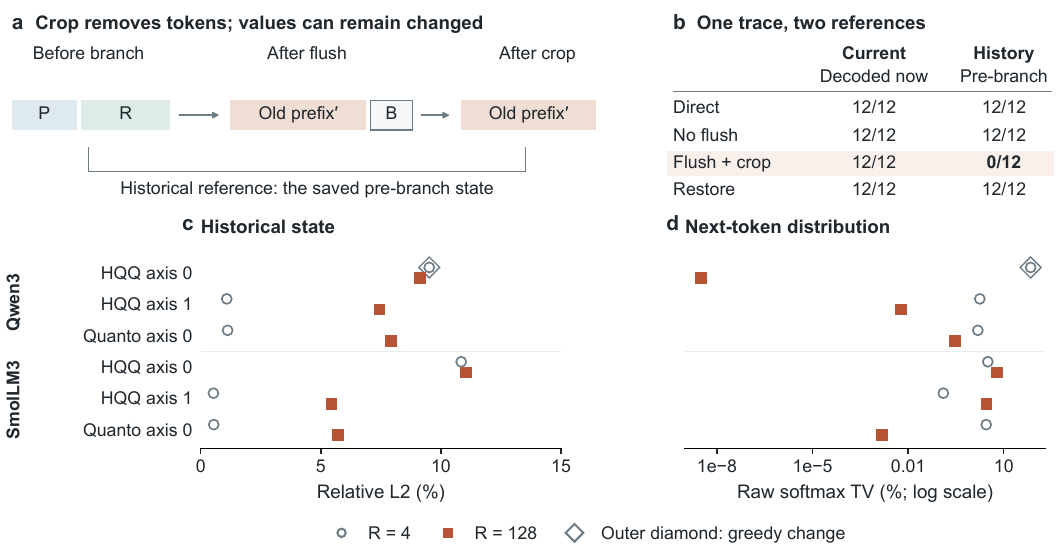}
\caption{Current-prefix correctness and historical restoration use different references. (a) P/R/B: packed/residual/branch; prime marks re-encoding. Cropping B need not restore earlier values; widths are schematic. (b) Four arms over six settings each at R=4 (initial residual one) and R=128 (125). Current checks pass when joint historical K/V and same-token logits fail. (c,d) All 24 measurements from two fixed traces, not a capacity-only comparison. Positive TV uses a log scale without pseudocounts; zero controls are in b. Outer diamonds mark one greedy-token change, preserving capacity markers, not an incorrect answer. Dense controls are exact in all twelve conditions; native removal is exact in eighteen paths without a retained-prefix flush. No task-failure rate is implied.}
\label{fig:rollback}
\end{figure*}

All 24 actual quantized consumers exactly matched their current-state references.
No-flush crop and checkpoint restoration also exactly matched direct
continuation. Yet all six flush--crop paths changed the retained history and
the next consumer's K/V and logits (Fig.~\ref{fig:rollback}). Their historical
KV drift ranged from 0.526\% to 10.843\%. Only one changed the greedy token;
the largest TV was 69.668\%, for Qwen3 HQQ axis zero. Cropping itself left
the packed payload unchanged and added no encoding.

A post-hoc regional audit separated each checkpoint's original packed prefix
(151 positions for Qwen3 and 390 for SmolLM3) from its one-token residual.
Every layer in all six settings had changed values in the old packed region.
Inserting only the original residual into the saved post-crop decoded arrays
left full-state relative-L2 errors of 0.110\%--10.829\%. The old packed region
accounted for 99.41\% and 99.73\% of total squared K/V error in the two HQQ
axis-zero settings, versus 0.96\%--10.24\% in the other settings. This locates
the precision loss relevant to exact historical restoration; it is an offline
array diagnostic on the same inputs, not a cache repair or another consumer
experiment.

A supplementary control, declared after those results, used the identical
four-token branch with DynamicCache initialized from the same checkpoint's
decoded K/V. All six crops and subsequent consumers exactly restored the
history reference. Unobserved direct consumers also matched their instrumented
counterparts. Saved-array and complete-state checks verified these controls. These are six
related settings on one question, not task-error frequencies. They establish
a concrete consumer-level contract boundary, not a new general rollback
principle, security evaluation or exact speculative-decoding implementation.

A separately frozen default-capacity check used the same six settings and
known prefixes with 125 accepted tokens from a saved no-maintenance
continuation. Capacity remained at its default 128: a one-token branch did not
flush, whereas a four-token branch did. All 24 current comparisons were exact; no-flush crop, complete
restoration and matched dense rollback recovered history. All six flush--crop
paths changed historical K/V and next-consumer logits: relative L2 was
5.448\%--11.046\%, and raw TV was \(3.14\times10^{-9}\%\)--6.029\%.
None changed the greedy token. Complete per-setting and post-hoc regional
results are retained in the replication material. The accepted segment used
one teacher-forced update; this boundary trace does not measure natural
rejection, workload frequency or answer quality.

A separately frozen Qwen3/HQQ axis-one witness accepted the same 125 tokens
through single-token updates. Native and adapted logits matched at every step.
The trace retained four exact current references and exact
no-flush, checkpoint, dense and observer controls. Flush--crop changed historical
K/V (7.445\% relative L2) and next logits (0.005925\% raw TV), without changing
the greedy token. This establishes sequential-update reachability in one
setting; the forced branch is not a public client's natural rejection path.

\subsection{Native removal in llama.cpp}
Unmodified llama.cpp~\cite{llamacpp2026pinned} used fixed-source official F16
GGUF conversion of Qwen3-1.7B, two existing prompts and native
F16/Q8\_0/Q4\_0 caches. CPU settings were two threads, context 256,
batch/microbatch limits 32, Flash Attention and native rotations. Seven paths
covered direct/repeated continuation, empty crop, append-one/four/eight then
crop and append-eight/checkpoint restoration; all consumed the same next
token. An independent reader checked codecs, metadata, all 28 layers' packed
K/V and full FP32 logits.

All eighteen append--crop paths (twelve quantized, six F16) retained the
original active state and direct-reference consumer logits; all controls
passed. Appends left earlier rows unchanged
without a retained-prefix flush. This within-runtime favorable contrast bounds
the failure by its path, not quantization alone. Different codecs and kernels
prevent treating the cross-engine comparison as a causal ablation; it does not
establish concurrent serving or universal correctness.

\subsection{Caller-visible updates in native MLX-LM}
\label{sec:mlx}
Alongside historical restoration, MLX-LM v0.32.2 promises in-place speculative-cache
updates~\cite{mlxlm0322}. A Swift report describes quantization leaving caller
references stale in a custom session~\cite{mlxSwiftCaller}. Here Python slices
the caller list into target/draft lists; quantization replaces their entries.
Synchronization propagates replacements:
\texttt{caller[:] = target + draft} on conversion; cleanup trims shared objects.
Three two-layer random Llama seeds with rejecting drafts expose twelve
4/8-bit prefill/verification conversion conditions: caller offsets seven/nine
versus internal eight after cleanup; nine controls match. Synchronization
preserves internal states and outputs. All 21 live next-model pairs yield identical logits from
the repaired caller target cache and the retained native internal target cache on
the same token. In this rejection matrix, stale callers change
logits but not argmax (Supplementary Fig.~S5). An unwrapped-target public-offset
regression fails natively and passes after repair.

At default threshold 5000, the same-seed
repeated-token fixture fails natively (target caller 5000, archived internal 4999).
Synchronization passes matched exhaustion/terminal-close checks with identical
outputs. Two upstream trim/save-load test bodies pass locally.
These tests cover single-step rejection/terminal-yield cleanup. The separate
\emph{Acceptance Follow-up} supplement tests accepted drafts and short repeated
verification: synchronization preserves internal computation, whereas
fixed-input stale-caller probes can change argmax. Arbitrary cancellation,
exceptions, concurrency and the full suite remain untested.

\section{RQ3: Decoding branches and observed outcomes}
\subsection{A public generation success can miss a rejection branch}
On a fixed random two-layer Qwen3 CPU model, native DynamicCache, HQQ axis one
and Quanto axis zero completed a constant-token prompt. An alternating prompt
reached \texttt{crop(-3)}: both quantized caches raised IndexError after one
forward, slicing an empty residual while leaving the packed prefix intact.
DynamicCache completed. This known failure class~\cite{transformers45510} is
not a pretrained-task failure. Our prefix-view extension completed all six
conditions; separate unobserved runs reproduced every output or exception.
These controls establish the scoped prefix contract, not a general decoding repair.

\subsection{Public sampling reaches a historical boundary}
The CPU study crossed prefix view/DynamicCache with lookup/AR on one known
88-token prompt: lookup three, ngram two, cap 160. Greedy and three sampled streams
(seeds 20261029--20261031, temperature 0.6, top-$p$ 0.95, top-$k$ 20) gave 32 calls:
sixteen exact observed/unobserved pairs and 2,128 observed forwards. A mismatched
warm entry remains excluded; unobserved forwards were not counted.

\begin{figure*}[t]
\centering
\includegraphics[width=\textwidth]{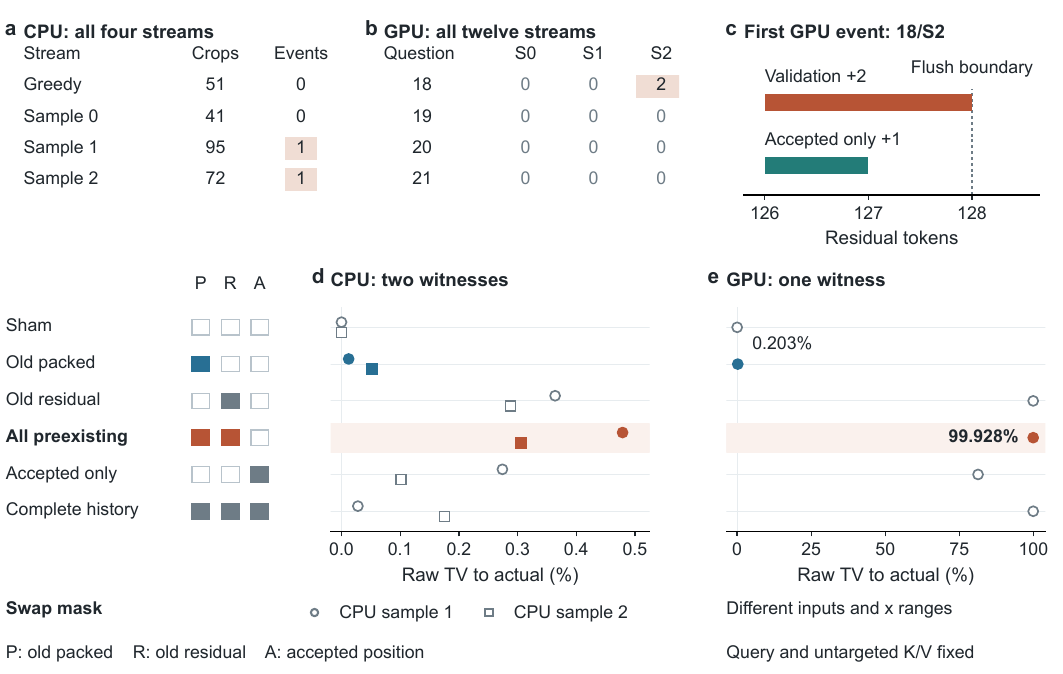}
\caption{Natural rejection and fixed-query history effects in the prefix-view extension. Qwen3-1.7B/HQQ four-bit axis one, group 64, R=128. (a,b) All CPU streams on one known prompt and all GPU streams from four new inputs/three seeds, including eleven zero-event streams. Crops counts negative-offset calls. (c) First 18/S2 event: residual 126 plus two validation inputs flushes; one accepted input does not. (d,e) Six arms fix query/shape; masks mark restored regions. CPU primary: first of four positions; GPU: one position. Axes use different linear ranges; distinct inputs do not isolate hardware effects. Sham/full-history reproduce actual/historical logits; untargeted K/V stays exact. CPU primary greedy stays; GPU all-preexisting removes the maximizer, old packed does not. Accepted-only donors include long/short arithmetic; mixed effects are non-additive, not repairs. All conditions cap (CPU 160/GPU 256); no production-rate, answer-harm or distributional-equality claim follows.}
\label{fig:public_history}
\end{figure*}

The frozen rule required an unforced rejection to trigger an existing-prefix
flush that accepted inputs alone would avoid. Only samples one/two qualified
(Fig.~\ref{fig:public_history}a): residual 124/125 plus four inputs crossed 128,
with one/two retained. Each event's nine calls passed exact physical-state,
crop-prefix, next-logit and current-dense controls. First-position raw TV to
history was 0.0279\%/0.1754\%. Dense long/short queries differed, preventing attribution
of the entire logit difference to history.

Following cache-state interventions~\cite{li2026phaselayouts,lorupStageReplay},
six post-hoc decoded-state arms fixed query and shape to separate old-history
effects from query geometry
(Fig.~\ref{fig:public_history}d). Sham/full-history endpoints and untargeted
values were exact. Preexisting-only first-position TV was
0.479\%/0.306\%; old 88-token packed-only TV was 0.0122\%/0.0516\%. These effects
are non-additive; mixed states are not repairs. All positions are retained,
the first is primary and its greedy token stays unchanged. All sixteen
conditions were capped. Same-seed lookup/AR does not couple draws or test
distributional equality.

A precommitted GPU check used the next four unused project inputs, source
rows 18--21, and three fixed seeds each. Two cache families, each with an
unobserved repeat, gave 48 lookup calls and 5,803 observed forwards at cap 256;
unobserved forwards were not measured. Model, cache and sampling settings and
the event rule matched the CPU study. Fig.~\ref{fig:public_history}b retains all twelve streams: eleven had no
event; question 18/seed 2 had two. Only its first event was diagnosed. A 52-token
visible packed prefix plus 126 residual tokens received two validation inputs; accepting
one alone would not flush. Fifteen diagnostic calls passed the same exact
controls, with independent NumPy validation.
Its actual next query had one position. Swapping only preexisting K/V changed
raw TV by 99.928\% and removed the actual maximizer from the new maximum set;
old packed alone changed TV by 0.203\% without removing it. The full historical
comparison had 99.930\% TV. All 24 conditions were capped; distinct CPU/GPU
inputs do not isolate hardware effects.
The accepted-only donor also contains long/short validation arithmetic:
its new position differs by 1.657\% relative L2 between matched dense paths.
It is not a pure quantization intervention; the old-history swaps leave
this accepted position fixed.

Post-hoc application of the original sampling filters (temperature 0.6,
top-$k$ 20, top-$p$ 0.95) to this GPU query's six saved arms assigns the observed
token probability 1 under sham and old-packed, 0 under old-residual,
all-preexisting and full-history, and 0.11634 under accepted-only.
Independent NumPy checks reproduce the support sets exactly. Thus, for this
fixed query, the old-history effect with the accepted position held fixed
survives sampling filters. This diagnostic establishes neither native-AR
equivalence nor accuracy or unconditional generation-distribution effects.

To check native-AR compatibility, a reconstruction of this known context held 52 packed and 126 forced
residual tokens. A 127-call replay matched 224 state tensors. Six continuations
gave exact native/view-AR compatibility; 18805 lay 5.8125 raw-logit units below
the native top-$k$ cutoff.
Twelve further calls used a full-validation-matched donor; all 56 returned
K/V were exact. Dense long/short query TV was
$5.19\times10^{-8}$, with both filtered supports $\{24\}$.
The six new fixed-query arms gave sham/old-packed probability 1,
old-residual/all-preexisting/full-history 0, and accepted-only 0.83952
for token 18805. Restoring old history removed its support with the actual
accepted position fixed and donor validation arithmetic matched. These remain
post-hoc, single-context checks; mixed effects are non-additive, not repairs.

An explicit client then tests a finite initial-prompt event
(Supplementary Fig.~S4). Clients A/B use native AR for 127 generated tokens;
B then enables lookup for two outputs. Three 129-call evaluations used
positive-weight forced prefixes and seven actual tail sampling rows (one
discarded). All 381 prefix vectors matched; native/view-AR logits matched
throughout. For fixed non-EOS prefix $h$ and the 129-token event $E=(h,28,18805)$,
the saved filtered vectors give unit conditional probability to both adopted
B tail tokens and exclude the final token under A. Fixed execution kernels
therefore give $P_A(E)=0$ and $P_B(E)=q>0$, with
$q=P_{\rm AR}(h)\approx1.41\times10^{-4}$. This is nominal categorical mass,
not seed probability or a certified numerical bound. The policy is not
always-on lookup.

On four further source-ordered inputs (22--25), three seeds each, all twelve
naturally sampled native-AR prefixes reached the fixed window. Eleven had no
candidate; one accepted its sole candidate. No rejection occurred, so this
cohort did not exercise the mechanism. All observed tails matched native AR;
this does not prove equal laws. The 1,595 calls include conditional branches,
not independent streams.

\subsection{Greedy stress does not establish answer degradation}
The 768 conditional greedy continuations were capped at 512 additional tokens.
All 128 lazy paths reproduced no maintenance exactly. Qwen3's 384 outputs all
hit the cap; SmolLM3 had 90 completed correct outputs and 294 caps. Same-shape
changed 122/128 paired Qwen3 trajectories and 124/128 SmolLM3 trajectories;
repeat-select changed 121/128 and 124/128. All comparable resolved maintenance
answers were correct. Caps and model-card restrictions limit this to
consistency stress; caps are not errors or evidence of general harmlessness.

\subsection{Sampling with model-card parameters gives mixed evidence}
The frozen two-question pilot used temperature 0.6 and top-$p$ 0.95 from the
model cards~\cite{qwen3pinnedcard,smolpinnedcard}. Top-$k$ was
20/50 for Qwen3/SmolLM3: the former is card-recommended, the latter the library
default retained in our configuration. Each model--question
pair shared a 64-token BF16 prefix and precommitted position uniforms for
token-ID inverse-CDF sampling, without replaying \texttt{torch.multinomial}.
The 4,096-token budget included the prefix. Ten paths per pair covered BF16
direct/re-prefilled continuation and both HQQ axes under four maintenance
paths (sixteen cycles). All forty outputs and two extra BF16 repeats remain.
All eight BF16 formal outputs completed with numeric answers (preset threshold
six); eight lazy and two BF16 repeat controls were exact. Independent checks
replayed 86,101 categorical draws and forty first-step distributions.

Of the forty correlated outputs, 25 had confirmed correct numbers, two had
confirmed incorrect numbers, and thirteen were capped. Both incorrect outputs
were the two BF16 paths for the same SmolLM3 question; its quantized paths
answered correctly. These observations must not be pooled as an accuracy
estimate over forty independent tasks. Supplementary Fig.~S1 retains all
paths, favorable outcomes and initial-quantization caps.

Qwen3 question 17/axis zero had a budgeted completion regression:
no maintenance/same-shape completed correctly at 2,470/1,497 tokens;
repeat-select hit 4,096. Question 16 already capped without maintenance,
preventing extra-effect attribution. No resolved maintenance answer changed
correct-to-wrong in this sixteen-cycle pilot; the one-cycle follow-up differs.

Local distribution differences also did not force a particular sampled choice.
For SmolLM3 question 17, axis zero, the first-step filtered TV between same-shape
and repeat-select was 0.2254, yet the fixed uniform selected the same token.
We separately report raw TV, filtered TV and the inverse-CDF coupling's local
disagreement probability. Neither is an answer-error probability, and sharing
uniforms in token-ID order need not minimize token disagreement.

\subsection{A single-cycle counterexample, with a failed expansion gate}
The mechanism follow-up completed all original settings. Eight full
no-maintenance replays exactly reproduced the earlier tokens, sampling traces,
text, first logits and parser outcomes. Together with sixteen new one-cycle
paths, the collection contained 24 outputs in 52,658 model calls. Fourteen
outputs had confirmed correct answers, one had a confirmed wrong answer, and
nine were capped. The repeated baselines are not new independent tasks.

One single-cycle condition changed a resolved answer relative to both controls.
For SmolLM3 question 17, axis zero, no maintenance and same-shape re-encoding
gave the correct 57,500. Repeat-select gave 59,800. The saved final answer used
52 weeks rather than the problem's 50 working weeks. All three paths shared
the prefix and precommitted random stream; the no-maintenance replay was exact.
This is a concrete conditional correctness counterexample, not an estimated
frequency of maintenance-induced errors.

\begin{table}[t]
\centering\small
\caption{Question 17, shared uniform stream. C: resolved correct; W: resolved wrong; U: capped/unresolved. Parentheses: total generated tokens. Cycle counts are related conditions, not independent tasks.}
\label{tab:single_cycle}
\renewcommand{\arraystretch}{1.08}
\begin{tabular*}{\columnwidth}{@{\extracolsep{\fill}}llrrr@{}}
\toprule
Model & Cycles & None & Same & Repeat\\
\midrule
Qwen3 & 1 & C (2470) & C (1830) & C (1603)\\
Qwen3 & 16 & C (2470) & C (1497) & U (4096)\\
\addlinespace[2pt]
SmolLM3 & 1 & C (677) & C (672) & W (931)\\
SmolLM3 & 16 & C (677) & C (647) & C (703)\\
\bottomrule
\end{tabular*}
\end{table}

The two cycle counts did not order outcome severity in these cases. The
SmolLM3 repeat-select answer was wrong after one cycle but correct after sixteen.
Qwen3 question 17, axis zero, completed correctly after one cycle, whereas its
sixteen-cycle repeat-select path was capped (Table~\ref{tab:single_cycle}). A
successful sixteen-cycle answer therefore did not certify the one-cycle answer
in the SmolLM3 case. These cases establish no accuracy ordering across
random streams.

The precommitted expansion rule (replication protocol
\texttt{maintenance\_single\_cycle\_followup\_v1}) required qualifying witnesses on two distinct
questions covering both models. Only one question and model qualified. The
gate failed, so we stopped expanding this candidate as a quality claim and
retained the counterexample and all favorable results. It remains development
on the same already observed two-question cohort, rather than independent
validation or evidence of universal harmlessness.

\section{Related work and scope of the contribution}
Compression/serving studies evaluate quantization, token selection and
quality/efficiency~\cite{liu2024kivi,li2024snapkv,tang2024quest,gao2025rethinking}.
SCBench evaluates cache
lifecycle and shared-context task performance~\cite{li2024scbench}.
We instead hold the encoded starting state fixed to isolate maintenance.

HQQ batch dependence and shared stopping~\cite{jwa2026sharedstopping} include
mixed groups (Appendix A) and a separation condition covering metadata/index
maps (Appendix B). Our equal-count/group controls test the layout boundary.
NVIDIA reports favorable NVFP4 repeated-QDQ
results~\cite{nvidia2026kvcompression}; BlockDialect's Appendix C identifies
requantization risks and buffers a high-precision tail until a block
fills~\cite{jang2025blockdialect}.

KVDiagnosis uses paired FullCache execution~\cite{qiu2026kvdiagnosis}; our
encoded-state reference excludes initial compression error. Public lazy
reorder~\cite{transformers45510,transformerscurrenttests}, state/logit~\cite{llamacpp2026pinned},
behavioral~\cite{ribeiro2020checklist} and diagnostic~\cite{exactkv2026} checks
are precedents. Phase-scoped layouts separate operator correspondence,
full-reference execution and state substitution~\cite{li2026phaselayouts};
stage replay tests live-state fidelity and causal cache sufficiency~\cite{lorupStageReplay}.
We apply these existing principles to encoded-cache maintenance.

QuantSpec defers quantization until verification using full-precision buffers
to avoid repeated QDQ during rollback~\cite{tiwari2025quantspec}, Section 4.3.
TransKV separates committed/speculative state~\cite{dongre2026transkv};
SubTernary uses a static scratchpad~\cite{yuan2026subternary}.
We measure acceptance consequences in fixed implementations
(Table~\ref{tab:decisions}).

Cache Engineering defines drop as preserving retained states and tracks
operation histories~\cite{fu2026cacheengineering}. Aborted but Not Forgotten
uses token-matched retained/rebuilt arms and DynamicCache
truncation~\cite{zhang2026aborted}, Section 4. Here crop removes rejected
positions without undoing lossy prefix rewrites; our historical reference is
the saved encoded pre-branch state. GRIEF fuzzes multi-request
behavior~\cite{zhao2026grief}; Patodiya separates trajectories from
quality~\cite{patodiya2026samerequest}. Swift's shared-ownership
change~\cite{mlxSwiftStorage} predates our Python contract measurements.
Our scope remains maintenance evidence, without new history-dependence
principles or security/concurrency evaluation.

\section{Interpretation and threats to validity}
\begin{table}[t]
\centering\small
\caption{Matched candidate-specific acceptance decisions; no defect-recall comparison.}
\label{tab:decisions}
\begin{tabularx}{\columnwidth}{@{}>{\raggedright\arraybackslash}p{.30\columnwidth}>{\raggedright\arraybackslash}X@{}}
\toprule
Check used & Matched observation and decision\\
\midrule
\textbf{Modeled proposal}\newline Equal budgets & Axis zero differs; group-preserving controls match. Equal counts do not certify preservation here.\\[3pt]
\textbf{Our v1}\newline Six public passes & \textbf{Reject v1.} Three composed conditions fail; corrected v2 passes all 96 tested conditions.\\[3pt]
\textbf{Our view, R128}\newline Current-prefix pass & Exact replay needs protected history: natural rejection changes it despite current-prefix passes.\\[3pt]
\textbf{Native MLX-LM}\newline Selected trim/save-load & Pass despite stale caller state; scoped synchronization passes the tested in-place checks.\\
\bottomrule
\end{tabularx}
\end{table}

\subsection{What changes a repair or test decision}
Table~\ref{tab:decisions} locates insufficient acceptance evidence: public passes
miss our repair's composition failure; current-prefix passes do not certify
history. MLX-LM tests caller visibility in parallel. Native removal, axis-one,
Quanto and lazy controls remain favorable; the unmerged proposal is not a
confirmed deployment defect.

\subsection{Numerical and instrumentation validity}
Independent float64 analysis uses saved arrays, which cannot exclude collection
errors. Repair references match batch shape and arithmetic. Public observers
have uninstrumented controls; failures remain recorded. Dense long/short queries
differed in BF16. Region swaps fix query/shape but create mixed states with
non-additive effects. Automatic SDPA kernel choice was not causally profiled.

Replication record \texttt{delayed\_lookup\_sampler\_v1} retains the failed warm
entry that replayed the prefix, excluded from the explicit event. Warm-prefix
input preparation has prior support~\cite{transformers42088}.

Equal counts do not isolate grouping, stopping or reduction order. The
original fixed-update/group audit used one known input; its L2 effects are non-additive.
Evaluation also depends on the endpoint: for example, CoLA-Flow separately
reports robotic task success, inference latency and trajectory smoothness~\cite{wu2026colaflow}.
Here, temperature, truncation and coupling affect local probabilities and tokens.
Changes in local probabilities or generated tokens alone do not establish
answer degradation.

\subsection{Construct and external validity}
Source-ordered greedy prefixes cannot estimate population rates or represent
recommended thinking trajectories. Sampling uses two questions, one stream
per model--question pair and baseline caps, precluding general accuracy claims.
Single-cycle checks probe mechanisms; sixteen cycles are stress, not frequency.

Numerical coverage is two Transformers families/two codecs; favorable
llama.cpp removal uses one model/two used prompts. Random fixtures are not
pretrained task failures. Reset checks use two short prompts/six output tokens
with an old cache module, not complete historical installations. Forced-stream
single-token checks cover Qwen3/HQQ axis one only. Natural rejection positives
remain in our prefix-view extension: one known CPU prompt and one qualifying
new GPU input out of four, retaining eleven zero-event streams. Public outputs
were capped; raw TV is not filtered-sampling error or answer harm. Production
rates, arbitrary compositions, offload, concurrency, defect recall, deployment
and performance remain unmeasured.

\section{Conclusion}
Identity drift persisted despite equal encodes. The original fixed-update and
group-preserving controls test its mechanism on one known input.

Flush/crop consumers matched current references despite changed history.
Checkpoint/dense rollback and native llama.cpp removal preserve tested history.
Public rejection in our extension and an explicit delayed client's 129-token
path establish conditional downstream effects; eleven zero-event GPU streams
and failed task-harm expansion bound these claims.

These measurements locate failures of acceptance evidence under stated
contracts. Historical references test exact pre-branch preservation; alongside
them, a scoped MLX-LM synchronization repair passes caller-state checks while
preserving native computations. Neither establishes general task harm.

\section*{Replication material and acknowledgments}
\textbf{Replication materials.} Protocols, code, figures and selected saved-vector
checks were collected for this study. Bulk raw states and failures are retained
locally. Fifteen GPU replay forwards matched 631 diagnostic tensors (118 copied
references excluded). Fresh all-setting reproduction remains untested.

\textbf{Generative-AI disclosure.} OpenAI Codex assisted research, code, analysis,
figures and writing throughout this work. Authors retain full responsibility.

\clearpage
\bibliographystyle{IEEEtran}
\IEEEtriggeratref{17}
\bibliography{bst-controls,references}
\end{document}